# Wearable Sensor-Based IoT-XAI Framework for Predicting Freezing of Gait in Parkinson's Disease

**Biplov Paneru[1],**

[1]Department of Electronics and Communication, Nepal Engineering College, Affiliated to Pokhara University,Bhaktapur, Nepal

Correspondence should be addressed to: Biplov Paneru, biplovp019402@nec.edu.np

***Abstract***— This research discusses the critical need for early detection and treatment for early prediction of Freezing of Gaits (FOG) utilizing a wearable sensor technology powered with LoRa communication. The system consisted of an Esp-32 microcontroller, in which the trained model is utilized utilizing the Micromlgen Python library. The research investigates accurate FOG classification based on pertinent clinical data by utilizing machine learning (ML) algorithms like Catboost, XGBoost, and Extra Tree classifiers. The XGBoost could classify with approximately 97% accuracy, along with 96% for the catboost and 90% for the Extra Trees Classifier model. The SHAP analysis interpretability shows that GYR SI degree is the most affecting factor in the prediction of the diseases. These results show the possibility of monitoring and identifying the affected person with tracking location on GPS and providing aid as an assistive technology for aiding the affected. The developed sensor-based technology has great potential for real-world problem solving in the field of healthcare and biomedical technology enhancements.

1. Introduction

Globally, Parkinson's disease (PD) / Freezing of Gaits is a major health concern that requires early detection for efficient management and intervention. Global crises including public health, climate change, and socioeconomic difficulties are being faced by the world today. In this regard, neurodegenerative illnesses like Parkinson's disease pose a substantial research challenge with broad ramifications. Globally, the prevalence of neurological disorders is rising, necessitating immediate attention and creative solutions. Parkinson's disease and other disorders are becoming more common as the population ages, creating difficult problems for healthcare systems and society at large. Multidisciplinary research initiatives that cross conventional boundaries and tackle complex problems are essential to comprehending and alleviating these crises. Healthcare costs are rapidly rising due to an aging population and a growing global population. Healthcare systems are undergoing a transition thanks to technologies like the Internet of Things, Edge of Things, and Cloud of Things, which enable health monitoring of individuals without the need for hospitalization [6]. In mid- and late-stage Parkinson's disease, freezing of gait (FOG), a severe gait abnormality that impairs movement and raises the risk of falls, is widespread. With the ultimate goal of preventing freezes or lessening their impact through gait monitoring and assistive technology, wearable sensors have been utilized to detect and forecast FOG [7]. Globally, Parkinson's disease (PD) is the second most common type of dementia. In recent years, wearable technology has proven helpful in both long-term surveillance and computer-aided diagnosis of Parkinson's disease. The key challenge is still how to accurately and efficiently use wearable technology to gauge the severity of Parkinson's disease [9]. Climate change and eating habits are just two of the numerous issues that healthy humans must deal with. To survive, the outcome needs to be conscious of the health condition. Health support services encounter issues such as overdiagnosis, data hazards, preventive errors, inaccurate patient information, and delayed implementation [10].

This study looks into the usefulness of machine learning (ML) algorithms in the classification of Parkinson's disease (PD) in response to this necessity. The research assesses the effectiveness of several algorithms, machine learning and deep learning method along with federated approach based on pertinent clinical data from frontiers publication [17]. This research not only addresses the urgent need for practical applications of ML in the field of Parkinson's disease diagnosis, but also advances predictive accuracy through the integration of traditional and deep learning algorithms.

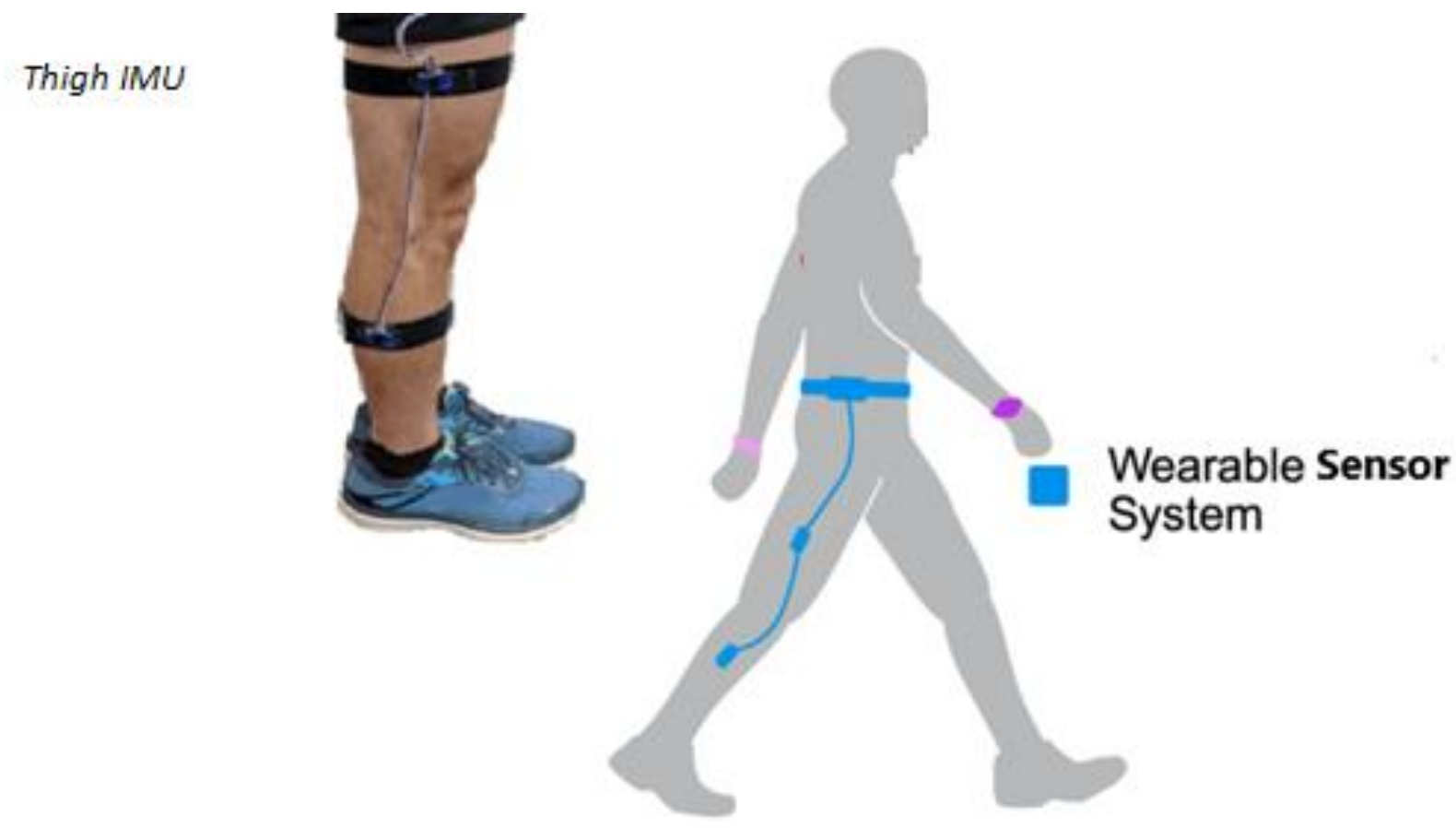

Figure 1 (a): Wearable sensor technology

1.1 Significance of research:

In 2017, the total economic burden of Parkinson's disease (PD) in the U.S. was $51.9 billion as shown in figure 1(c) as per report published on [19], consisting of:

- Direct medical costs: $25.4 billion
- Indirect and non-medical costs: $26.5 billion, including:
    i. Indirect costs ($14.2B): lost earnings from premature death, reduced employment, absenteeism, presenteeism, and volunteer productivity loss
    ii. Non-medical costs ($7.5B): paid care, home/vehicle modifications, other related expenses
    iii. Disability income: $4.8 billion paid to individuals with PD

Parkinson's disease (PD) is a growing global health concern, especially as populations age. It primarily affects movement, causing symptoms like tremors, stiffness, and slowed mobility. With over 1.5 billion people expected to be over age 65 by 2050, the number of PD cases is projected to rise significantly. In the U.S. alone, nearly one million people live with PD, a number expected to reach 1.2 million by 2030. While most cases occur in older adults, about 4% of those diagnosed are under 50, and men are 1.5 times more likely to be affected than women. The Parkinson's disease diagnosis and treatment market is projected to grow from USD 7.5 billion in 2025 to USD 16.14 billion by 2034 [20].

Early detection of Parkinson's is crucial for managing symptoms and improving patient outcomes. Artificial Intelligence (AI) holds great potential in predicting early signs of PD, especially by analyzing gait (walking patterns) as shown in figure 1(a) through sensors data, which often changes subtly in the early stages. AI-powered tools can detect these changes through motion sensors and video analysis, allowing for earlier diagnosis and timely intervention. This can significantly enhance quality of life and reduce the burden on healthcare systems by starting treatment before symptoms become severe.

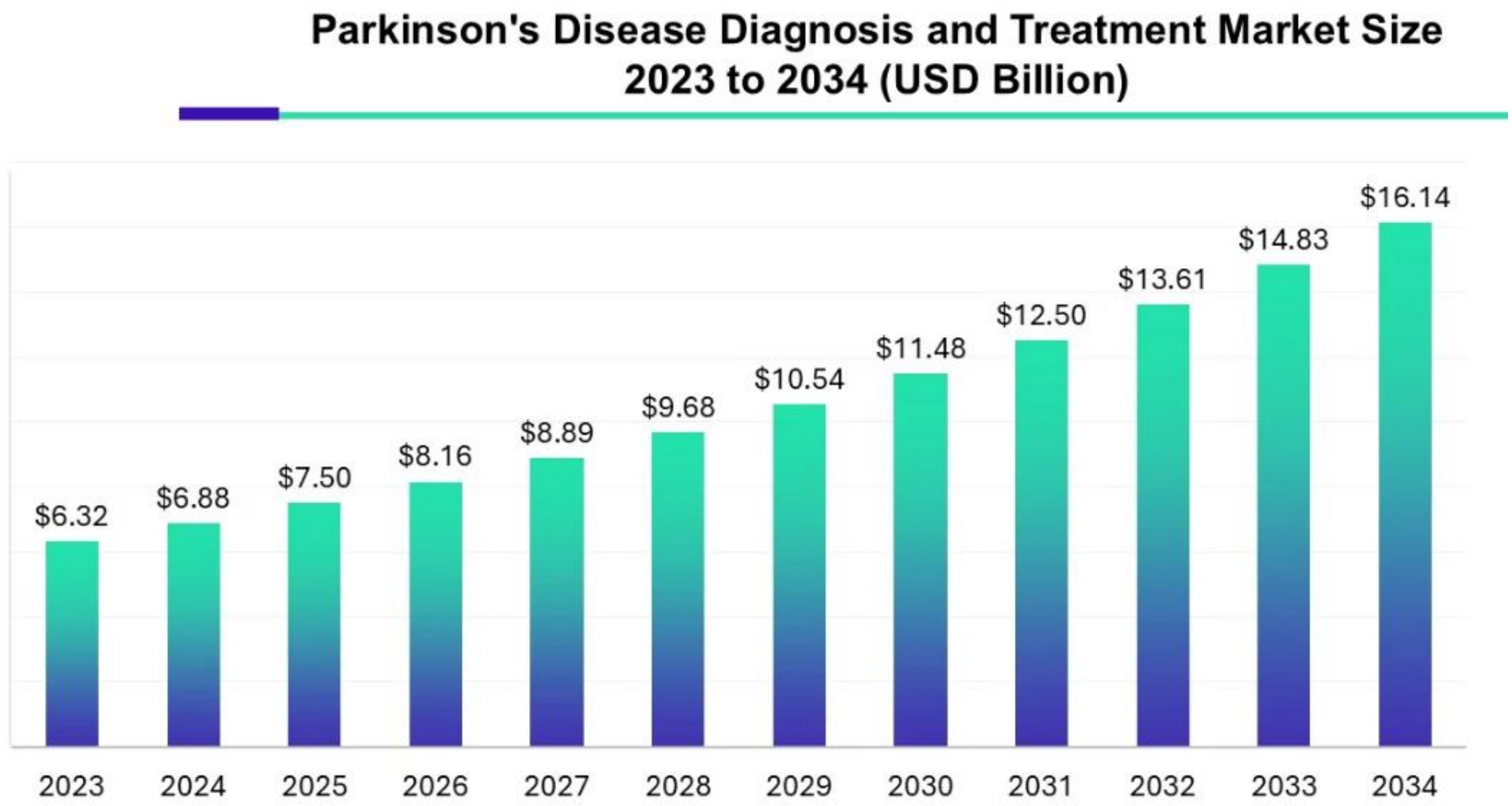

Fig 1(b): Global estimate of Parkinson's market forecast

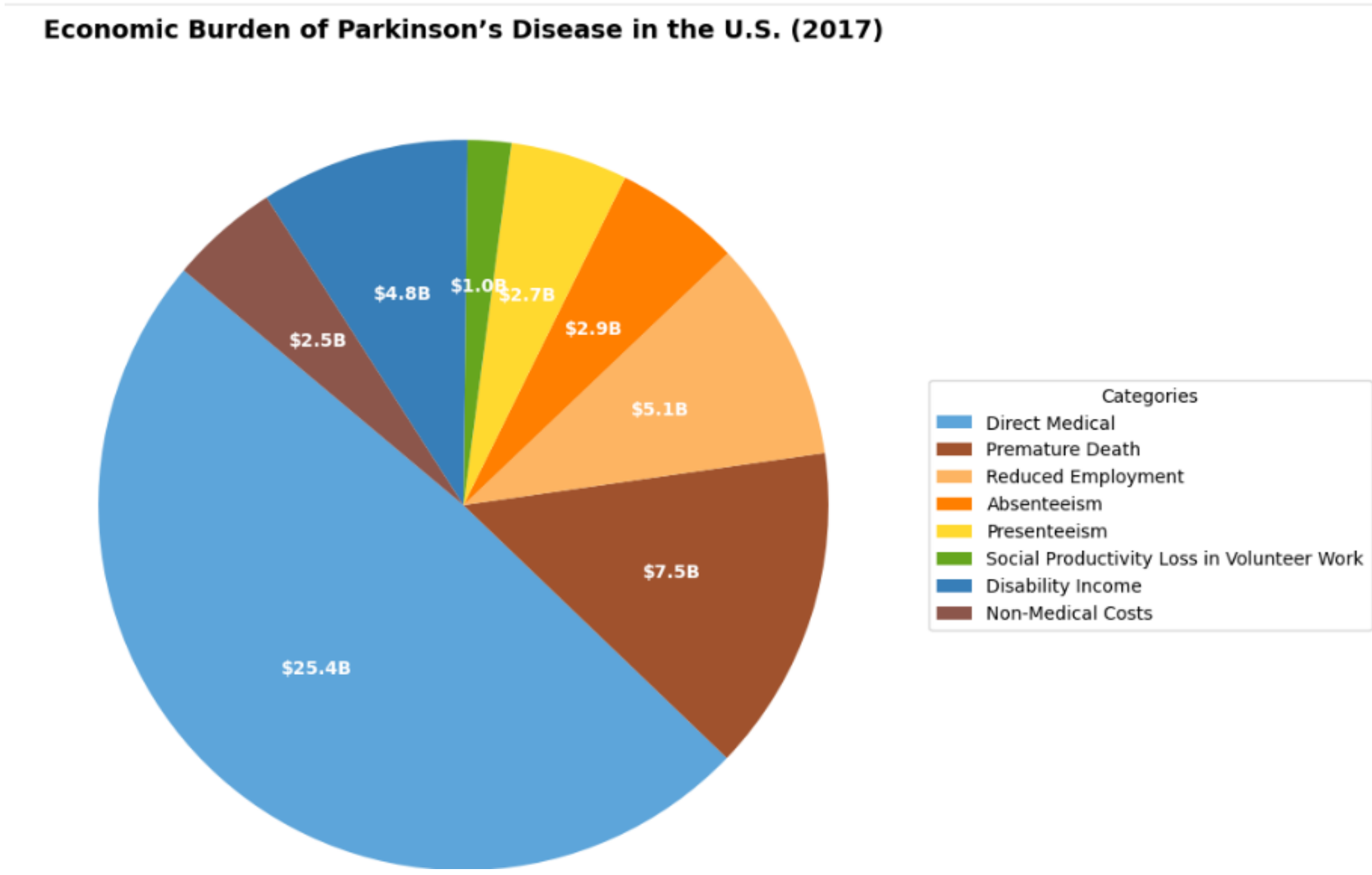

Fig 1 (c): Economic burden caused by Parkinson's disease in US [19]

## 1.1 Literature Review

The application of machine learning (ML) techniques for the diagnosis and progression monitoring of Parkinson's Disease (PD) has received significant attention in recent years. Various studies have integrated supervised learning, deep learning, wearable technology, and signal processing to enhance diagnostic accuracy and real-time monitoring.

1. *Supervised Machine Learning in PD Diagnosis*

Alshammri et al. (2023) implemented a hybrid approach using GridSearchCV for hyperparameter tuning and SMOTE for handling class imbalance, achieving 99% F1-score with MLP and 97% with SVM, showing promise for early PD diagnosis through structured ML workflows [1]. Similarly, Nithya et al. (2021) applied a hybrid of SVM and Random Forest on MRI images aligned using image registration, highlighting enhanced specificity and diagnostic accuracy [3].

*2. Ensemble and Deep Learning Approaches*

Ezhilin Freeda et al. (2022) explored early prediction using speech-based biomarkers and ensemble classifiers such as XGBoost and Random Forest, which outperformed traditional decision trees [4]. Schmid et al. (2022) proposed a combination of Rotation Forest and Random Forest on voice data from the UCI dataset to classify severity levels [5].

Chaterjee et al. (2023) introduced the PDD-ET model, which significantly outperformed other models, including CNNs, LSTMs, GRUs, and DNNs across multiple metrics by margins ranging from 16% to 45%, illustrating the robustness of their ensemble approach [2].

*3. Sensor-Based Monitoring and Deep Learning*

Uddin et al. (2021) designed a sensor-based RNN system for activity prediction using wearable medical sensors like ECG, accelerometers, and gyroscopes. The model, deployed on edge devices, outperformed traditional methods in speed and accuracy [6]. Likewise, Mishra et al. (2022) used wearable sensors to collect data from 39 Friedreich's Ataxia (FRDA) patients and developed ML models using sensor-derived, biological, and demographic features to predict disease severity [8].

Yue et al. (2024) proposed a Latent Dirichlet Allocation (LDA)-based architecture to capture activity-derived latent features, achieving a classification accuracy of 73.48% across PD severity levels in free-living settings [9].

*4. FOG Detection and Gait Analysis*

Freezing of Gait (FOG) remains a significant challenge in PD management. Pardoel et al. (2019) reviewed 74 studies, finding that modern ML models with transfer learning and semi-supervised learning offer improved person-specific FOG detection, despite challenges like limited datasets [7].

Deep learning-based approaches such as DeepFoG by Bikias et al. (2021) used IMU sensor data and achieved up to 90% specificity in FOG detection via LOSO and 10-fold cross-validation [12]. Similarly, Borzì et al. (2022) implemented a multi-head CNN on inertial sensor data from 118 PD patients, predicting FOG onset 3.1 seconds earlier with high specificity (>88%) [13].

Salomon et al. (2022) conducted a global ML competition involving 1,379 teams to develop wearable-based FOG detection models. The winning solutions showed strong correlations to gold-standard datasets and uncovered previously unnoticed real-world FOG patterns [14].

Shi et al. (2023) applied Continuous Wavelet Transform and Bayesian optimization to train CNNs on IMU signals, achieving an impressive 91.5% F1 score in subject-independent FOG detection [15].

*5. IoT and Intervention-Based Systems*

Pradhan et al. (2021) integrated IoT-based data collection with a Boltzmann-trained AI feedback system, achieving a 97.4% prediction rate for early disease detection [10]. Furthermore, a recent study by Kim et al. (2024) introduced soft robotic garments to prevent FOG by enhancing hip flexion. This led to measurable improvements in walking distance (+55%) and gait quality in a six-month case study [16].

The review of the literature identifies a number of interesting strategies and methods in the field of machine learning-based Parkinson's disease prediction. It also highlights important research gaps that need to be looked into further. This proposed method applies the machine learning models on a new dataset from frontiers publication and utilizes explainable AI with traditional models in order to integrate with microcontrollers to deliver the victim aid notifications and alert early via IoT system.

## 2. Materials and methods

The study's dataset [21] includes the IMU data collected through various subjects that are important for diagnosing Gait disease. To improve the quality of the input for the ML models, preprocessing steps include feature selection, normalization, and data cleaning. The performance of the model is then assessed by dividing the dataset into training and testing sets. Using cross-validation, the selected machine learning algorithms are put into practice and their hyperparameters are optimized. The table 1 shows hardware components for IoT integration to the system.

Table 1: Hardware description

| Hardware component | Function |
|---|---|
| Esp-32 | Main connectivity-based microcontroller (supports wi-fi) |
| LoRa | Long range transmission module |
| IMU sensor (MPU-6050) | Getting 6 axis accelerometer / gyro data |
| GPS module | Getting location (latitude and longitude) |
| 5v battery | Power supply main source |

**Table 2:** Dataset example

| Time [s] | ACC ML [g] | ACC AP [g] | ACC SI [g] | GYR ML [deg/s] | GYR AP [deg/s] | GYR SI [deg/s] | Freezing event [flag] |
|---|---|---|---|---|---|---|---|
| 1.359375 | 0.105983 | -0.32385 | 0.838768 | 8.897047 | -16.8301 | 33.93851 | 0 |
| 1.367188 | 0.142105 | -0.18689 | 0.883557 | 9.206729 | -19.46 | 34.4718 | 0 |
| 1.375 | 0.101335 | -0.18438 | 0.924018 | 5.610897 | -21.3965 | 29.59217 | 0 |
| 1.382813 | 0.050564 | -0.16351 | 0.947961 | 0.715407 | -22.4346 | 20.82936 | 1 |
| 1.390625 | -0.07304 | -0.21369 | 0.879053 | -1.75361 | -22.0239 | 17.14693 | 1 |
| 1.398438 | -0.04873 | -0.2424 | 0.838982 | -5.22388 | -25.6008 | -6.74861 | 1 |
| 1.40625 | 0.047031 | -0.37857 | 0.847576 | -9.90098 | -27.0137 | -21.2456 | 1 |

**2.1 System concept:**

As, seen in the figure 3 above the IMU sensor data along with GPS and LoRa module data can be received using a microcontroller Esp-32 then, the data can be live monitored using IoT technology and the caretaker can be messaged or notified easily depending upon the emergency situation need. The power source is applied with a 5v rechargable Li-on battery. The Li-on battery can be recharged per need and use of IMU is an approach for detecting gait with Machine Learning model integrated to microcontroller via Micromlgen python library. Figure 4 shows proposed system connection.

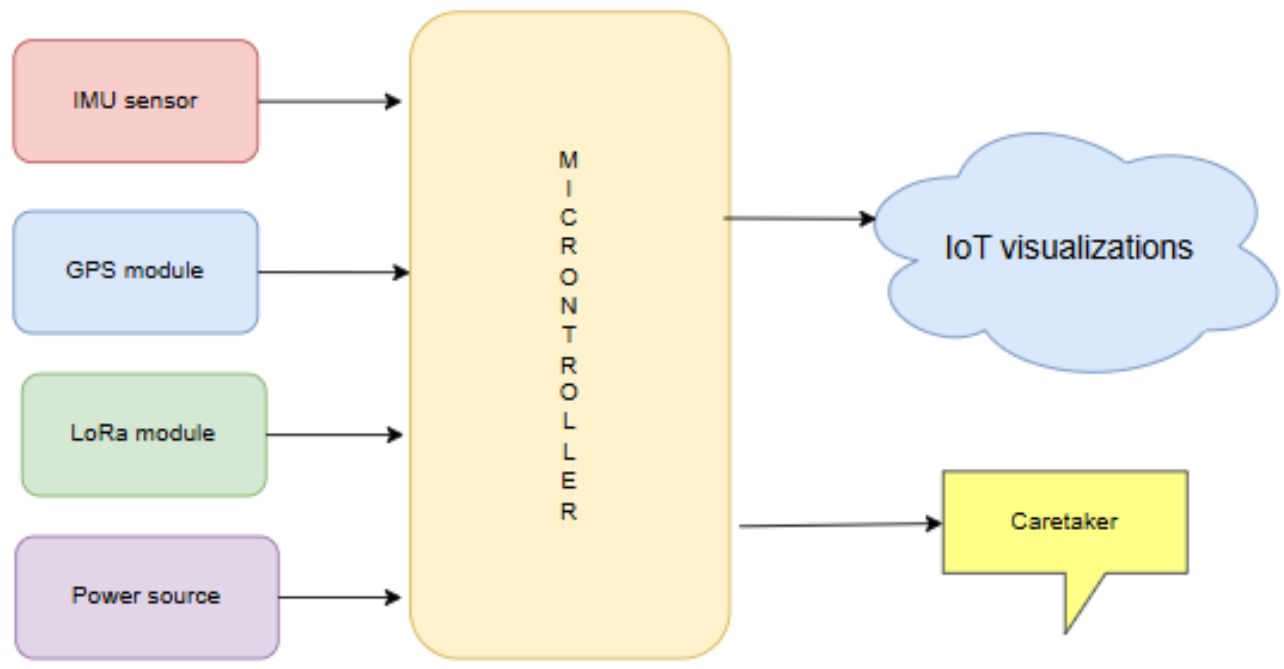

Figure 3: System block diagram

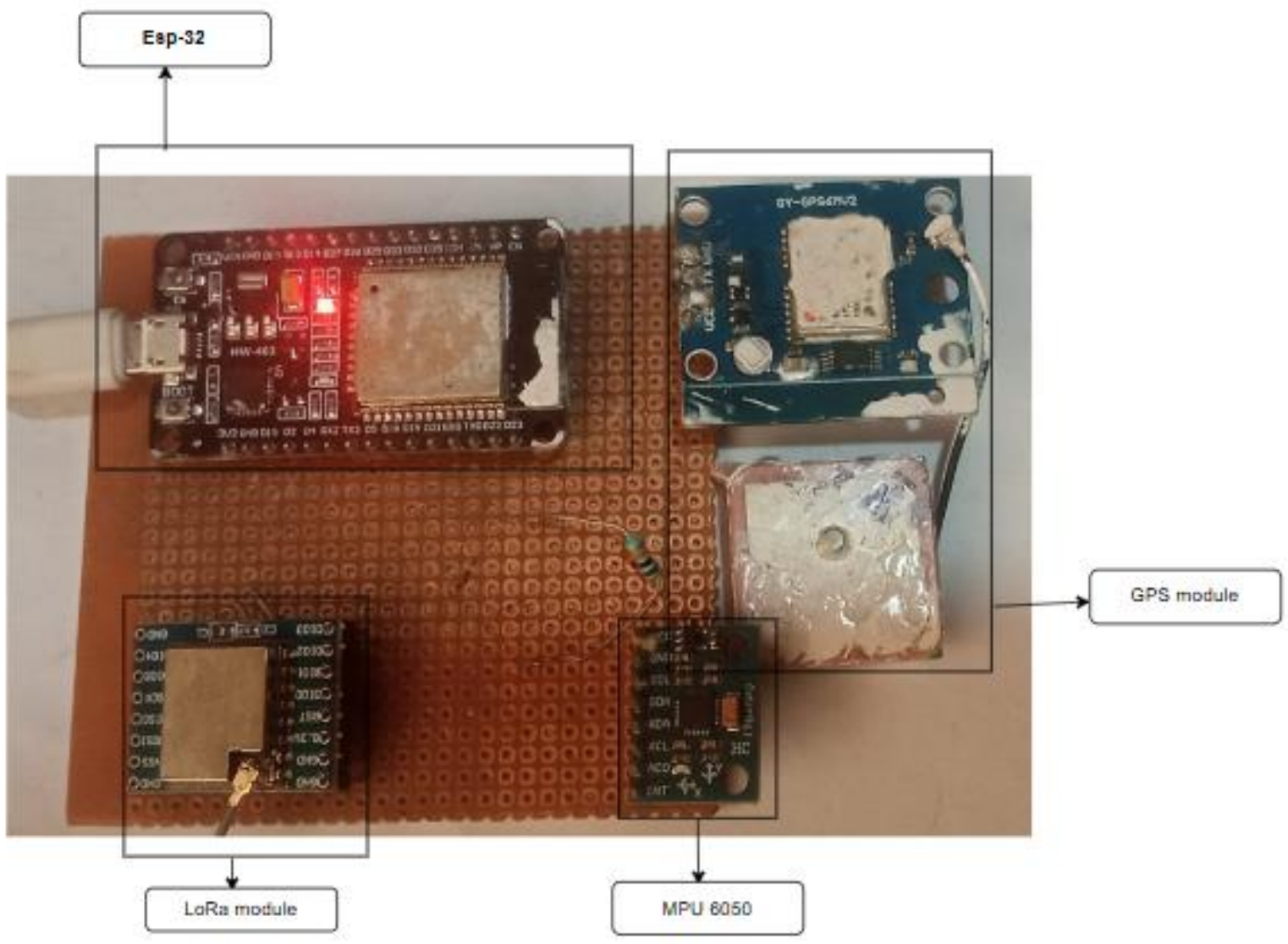

Figure 4: proposed system connection

### *a. Model selection*

Along with various ML algorithms we utilized different deep learning algorithms too. Two state-of-the-art machine learning models, Long Short-Term Memory (LSTM) and Multi-Layer Perceptron (MLP), were deliberately selected in order to achieve robust and accurate Parkinson's disease classification. Recurrent neural networks (RNNs) of the LSTM type are particularly good at identifying temporal dependencies in sequential data, which makes them a good choice for examining time-series elements in the clinical dataset. However, because MLPs are feedforward neural networks, they can handle a wide range of complex non-linear relationships in datasets with lots of features. The choice to combine both XGBoost and Catboost is based on the necessity of utilizing each class's advantages in managing complex patterns and temporal dynamics in clinical features.

### *b. Dataset selection*

The dataset collected from the Frontier's publication source consists up of the IMU sensor data regarding Freezing of Gait 2 conditions that are Gait, and No-Gait. The use of ML model in the micromlgen python library especially needs a lightweight model development. So, we are dependent to a balanced volume dataset for developing 3 different models for purpose of Gait Prediction using IMU sensors data. The dataset combined through excel worksheet from SUB 01_1, and SUB 2_05 [17]. The dataset volume content can be seen in figure 5.

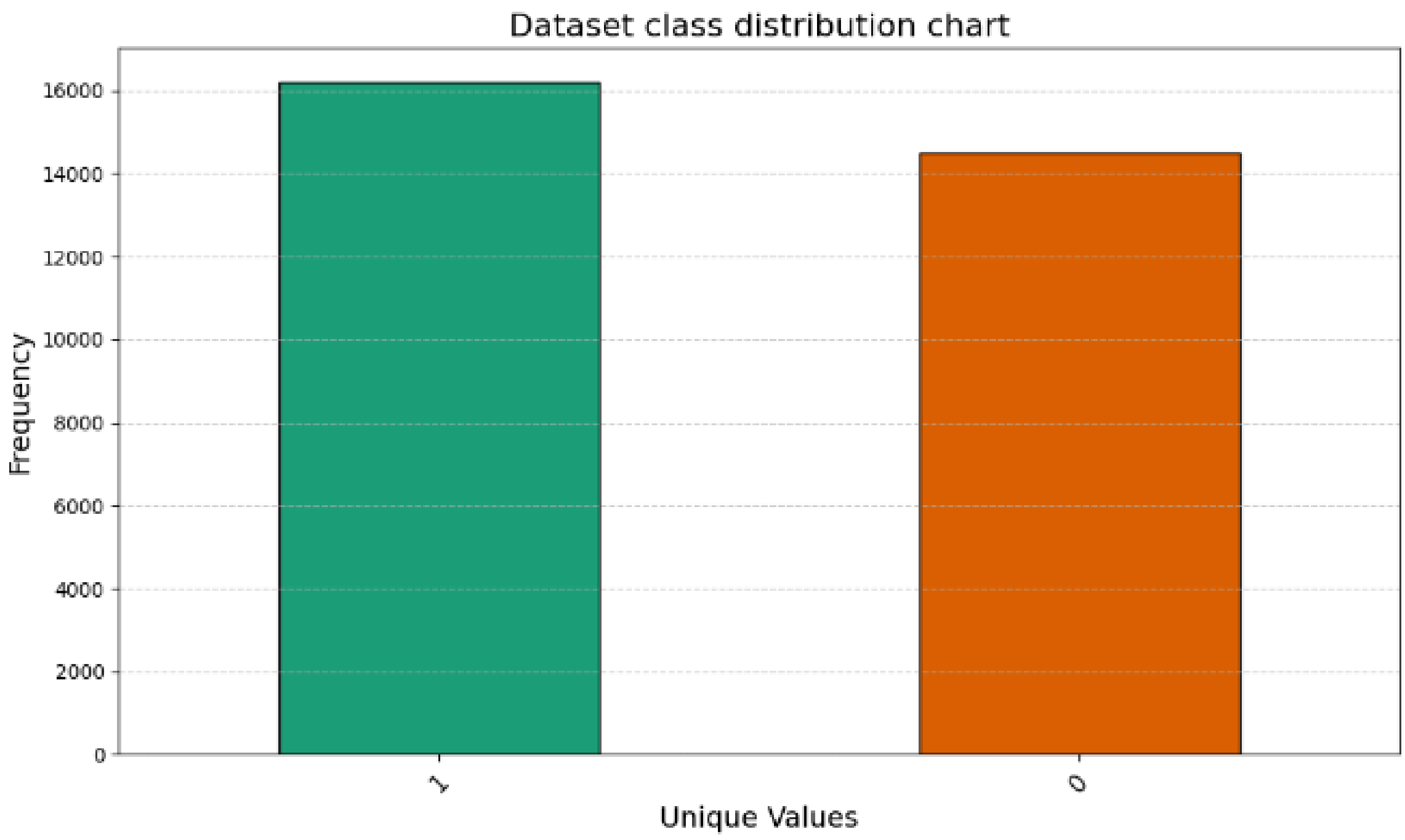

Figure 5: Data classes division

***c. Model's proposed***

1. Extreme Gradient Boosting, or XGBoost, is a well-known and potent ensemble learning algorithm. In order to produce a reliable and accurate predictive model, it integrates the predictions of several weak models, usually decision trees. XGBoost is particularly good at managing intricate relationships in the data, reducing overfitting, and offering resilience against anomalies. Its efficacy is demonstrated by its popularity in data science competitions, which makes it a useful tool for tasks where predictive accuracy is critical, such as Parkinson's disease classification.
2. Catboost: CatBoost is a gradient boosting algorithm created by Yandex with the goal of effectively handling category data. Without the requirement for explicit encoding, like one-hot encoding, it automatically processes categorical features, which can speed up preprocessing and enhance model performance. High accuracy, speed, and the ability to avoid overfitting using a variety of regularization strategies are all attributes of CatBoost. Regression and classification tasks frequently use it, and it works well with datasets that contain categorical variables and intricate relationships.\

3. Nested cross validation: Two levels of cross-validation are used in nested cross-validation, a model evaluation method that reduces overfitting and yields a more accurate estimate of model performance. The dataset is divided into training and testing subsets by the outer loop, and the test set is used to assess the model. The inner loop optimizes the model's hyperparameters by cross-validating the training data of each outer fold.

**Evaluation metrics for model**

i. **Accuracy**: The ratio of correctly predicted instances to the total instances.

$$\text{Accuracy} = \frac{TP+TN}{(FP+FN+TP+TP)} \qquad (3).$$

ii. **Precision**: The ratio of correctly predicted positive instances to all predicted positives.

$$\text{Precision}=\frac{TP}{(FP+TP)} \qquad (4)$$

iii. **Recall**: The ratio of correctly predicted positive instances to all actual positives.

$$\text{Recall} = \frac{TP}{(FN+TP)} \qquad (5)$$

iv. **F1-Score**: The harmonic mean of precision and recall, balancing both metrics.

$$\text{F1 Score}=2 \times \frac{Precision*Recall}{Precision+Recall} \qquad (6)$$

## 3. Results and discussions:

The XGBoost model can be seen to outperform other models with 99.95% training accuracy and 96.78% testing accuracy. The model XGBoost is further nested cross validated for performance measure. The catboost model also has shown a great result with 99.85% training accuracy and 99.24% testing accuracy and finally, the extra trees classifier model obtained 94.65% training and 90.33% testing accuracy. These metrics are helpful for evaluating these model performances and their usability with real-time sensors integrating for the purpose of detecting and diagnosing Freezing of Gait early.

Table.3. Accuracies results

| Model | Training | Testing |
|---|---|---|
| XGBoost | 99.95% | 96.78% |
| Catboost | 99.85% | 96.24% |
| Extra Trees | 94.65% | 90.33% |

However, the XGBoost model demonstrated exceptional accuracy, scoring 78% on the testing set and 99% on the training set. This highlights how ensemble learning can manage complex relationships in the data, but the small decline in testing accuracy suggests that further investigation and model modifications may be necessary.

The training and testing accuracy of the Gaussian Naive Bayes algorithm were 84% and 79%, respectively. This indicates that the algorithm can effectively simulate the probabilistic relationships found in the clinical data, offering an accuracy level appropriate for the task at hand.

**3.1 XGBoost:**

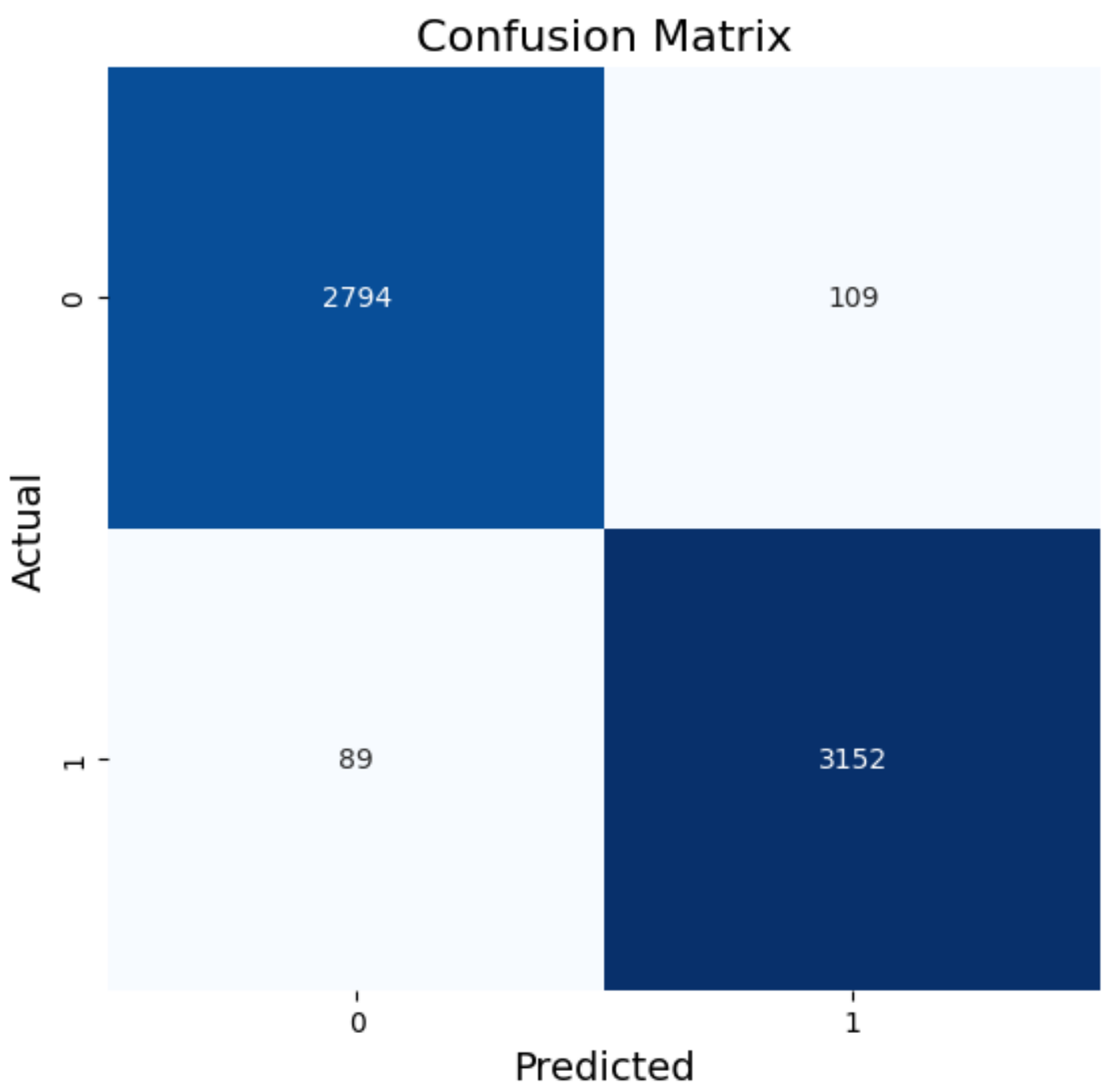

Figure 6: XGBoost confusion matrix

Out of the 2903 instances in the total cases for category 'No-FOG' meaning class '0', the model correctly, as shown in Fig. 5, predicted 109 cases of gait (true positives) and 2794 cases of no-gait, according to an analysis of the confusion matrix, which shows the model's predictions. Nonetheless, there were five cases (false negatives) in which the model predicted a condition other than Parkinson's when the true condition was Parkinson's, and five cases (false positives) in which the model predicted a condition other than Parkinson's when the true condition was Parkinson's. Out of 3241 instances, 3152 were predicted correctly, and 89 instances were wrongly predicted as class '1' meaning 3152 instances were correctly predicted as gait.

Table 4: 10- Fold nested cross validation results:

| Fold | Test Accuracy (%) |
|---|---|
| 1 | 96.74 |
| 2 | 96.48 |
| 3 | 96.74 |
| 4 | 96.32 |
| 5 | 96.19 |
| 6 | 96.22 |
| 7 | 96.09 |
| 8 | 97.01 |
| 9 | 96.48 |
| 10 | 96.06 |
| **Mean Accuracy** | **96.44** |
| **Standard Deviation** | **0.30** |

### 3.2 Cross validation Results

The outcomes of the five-fold cross-validation as shown in fig.5. conducted on the Parkinson's disease dataset offer significant insights into the predictive model's performance. The model's accuracy in each fold was 96%, meaning that 96% of the time it correctly identified cases as either FOG is occurring or not and a very low standard deviation of value is obtained approximately 0.30 only.

Out of the 2903 instances in the total cases for category 'No-FOG' meaning class '0', the model correctly as shown in fig.5. predicted 120 cases of non-Parkinson's (true negatives) and 2783 cases of Parkinson's (true positives), according to an analysis of the confusion matrix, which shows the model's predictions. Nonetheless, there were five cases (false negatives) in which the model predicted a condition other than Parkinson's when the true condition was Parkinson's, and five cases (false positives) in which the model predicted a condition other than Parkinson's when the true condition was Parkinson's. Out of 3241 instances 3130 were predicted correctly and 111 instances were wrongly predicted as class '1'.

### 3.3 Catboost:

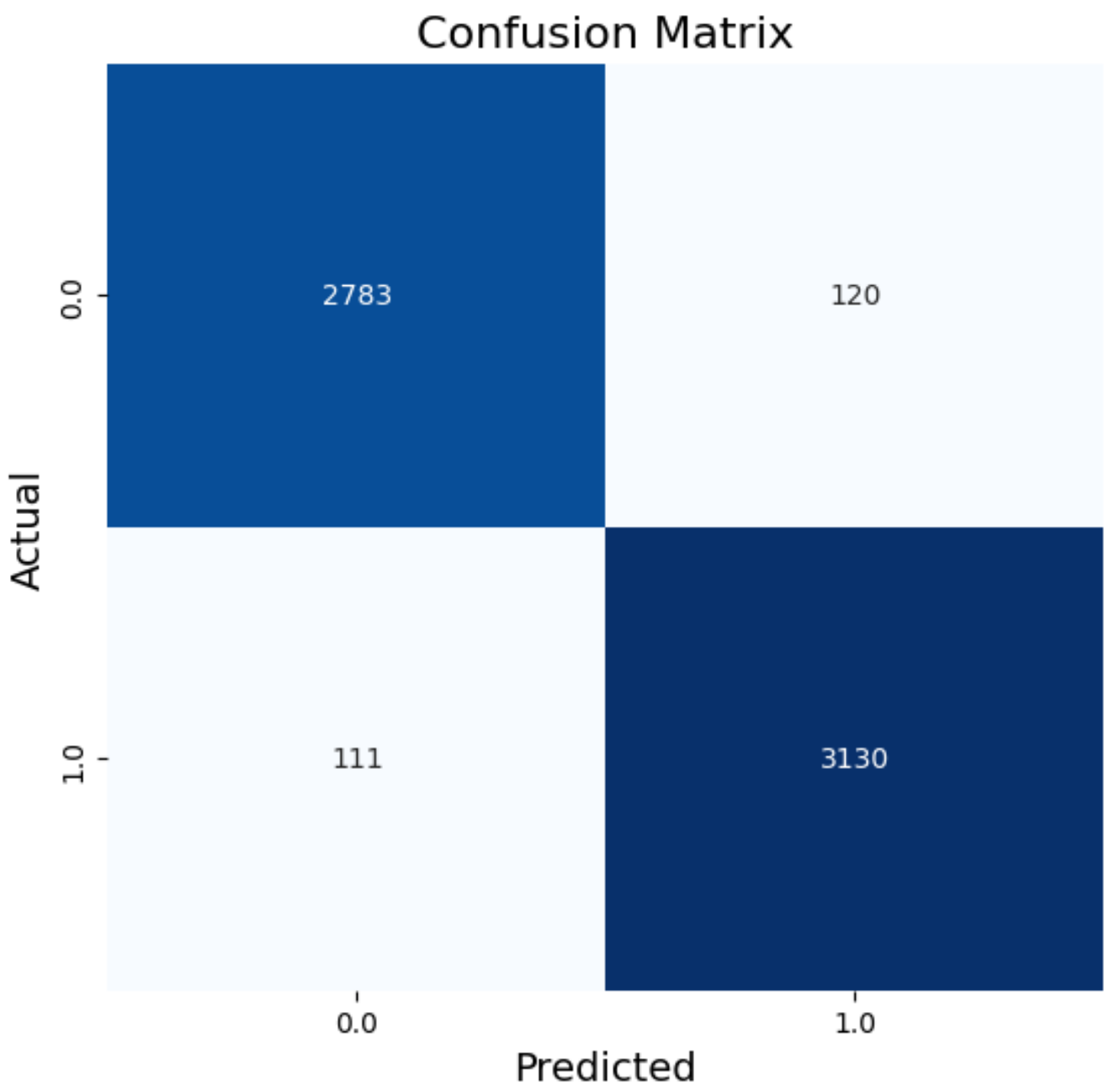

Figure 7: Catboost confusion matrix

Out of the 2903 instances in the total cases for category ‘No-FOG’ meaning class ‘0’, the model correctly as shown in fig.7. predicted 120 cases of non-Parkinson's (true negatives) and 2783 cases of Parkinson's (true positives), according to an analysis of the confusion matrix, which shows the model's predictions. Nonetheless, there were five cases (false negatives) in which the model predicted a condition other than Parkinson's when the true condition was Parkinson's, and five cases (false positives) in which the model predicted a condition other than Parkinson's when the true condition was Parkinson's. Out of 3241 instances 3130 were predicted correctly and 111 instances were wrongly predicted as class ‘1’.

### 3.4 Extra Trees classifier

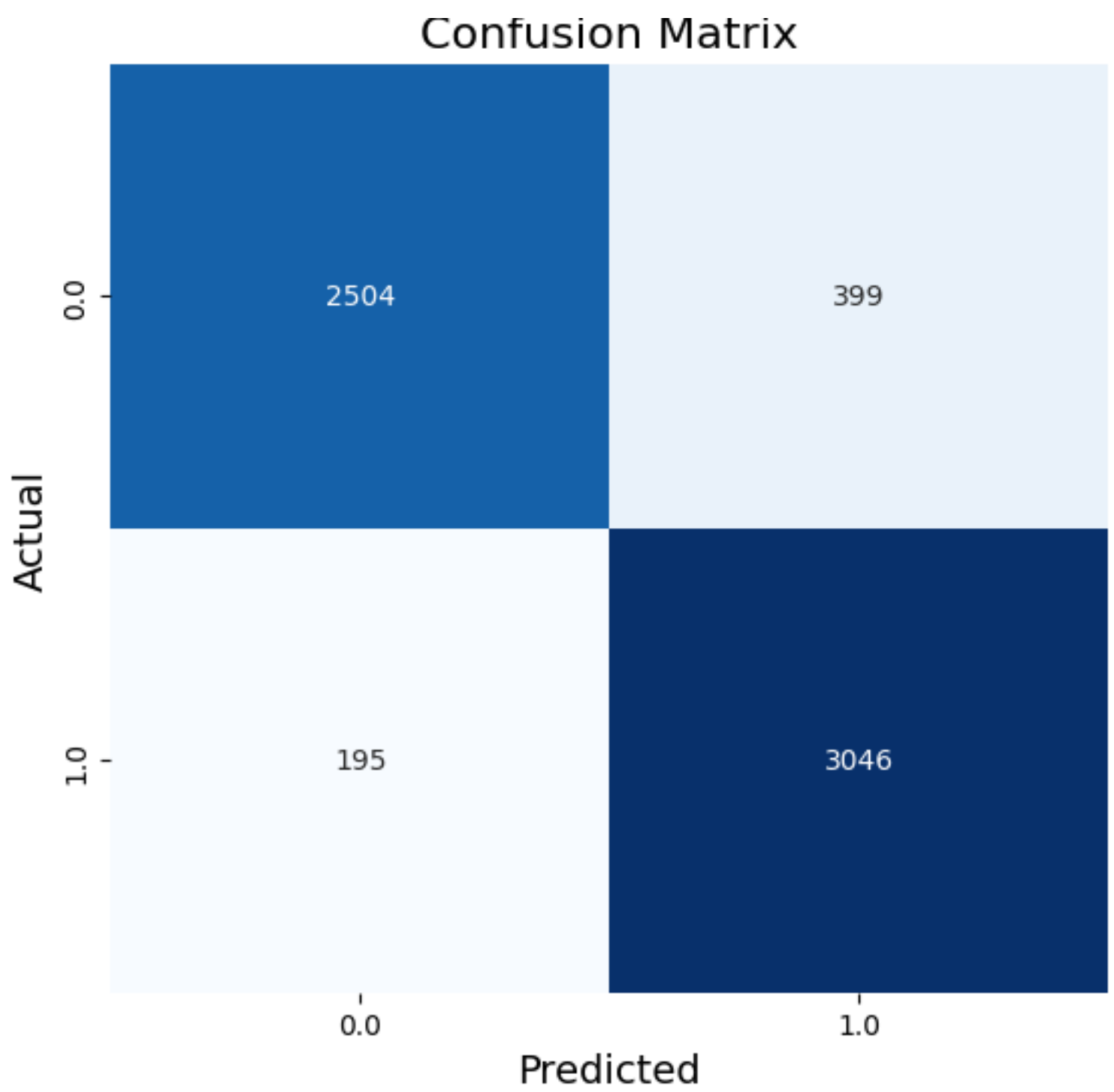

**Figure 8:** Extra Trees confusion matrix

The confusion matrix on XGBoost model shows that 2504 instances were correctly predicted as no gait and similarly, 3046 were predicted as '1' meaning gait was predicted. The Extra Trees classifier light weight model too shows an exemplary result for potential deployment in the microcontroller for IMU data-based predicting gait.

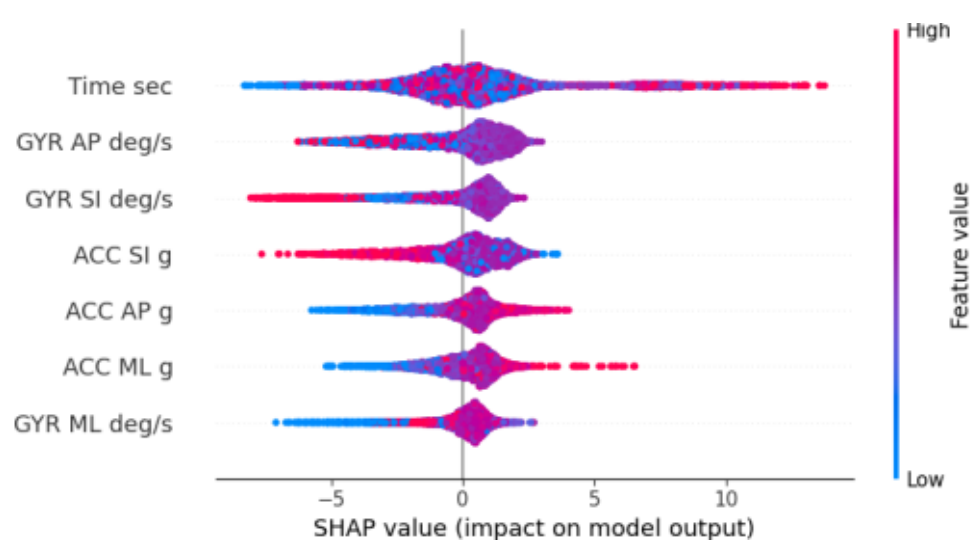

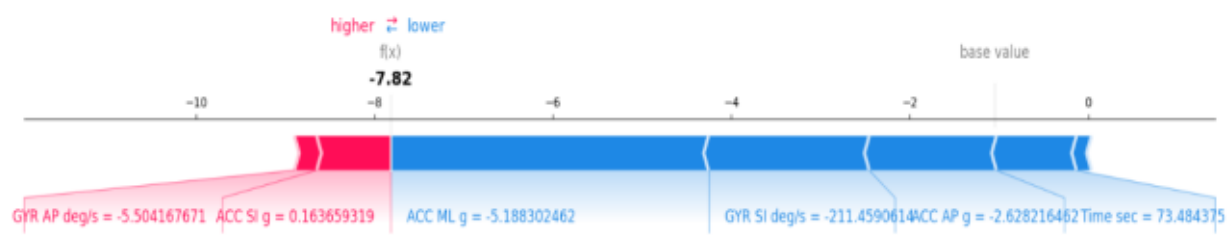

Fig 9 (a): Affecting feature analysis plot  Fig 9 (b): Model SHAP analysis on XGBoost

The SHAP summary plot in figure 9 visualizes the impact of various features from Inertial Measurement Unit (IMU) data on a machine learning model's prediction of Freezing of Gait (FOG) events. Each feature's contribution to the model output is depicted along the x-axis as SHAP values, representing the magnitude and direction of influence. Features with high SHAP values strongly affect the model's predictions and GYR SI deg/s" (red from the IMU is seem to be affecting the model performance mostly than other metrics showcasing its contribution in Gait or no gait condition classification.

The features include time-related data ("Time sec"), gyroscope readings ("GYR AP deg/s", "GYR SI deg/s", and "GYR ML deg/s"), and accelerometer readings ("ACC SI g", "ACC AP g", and "ACC ML g"). Each point corresponds to a data sample, and the color indicates the feature value, transitioning from low (blue) to high (red). For example, high values of "GYR SI deg/s" (red) generally increase the SHAP value and positively influence the prediction of FOG. Conversely, features like "ACC ML g" show diverse impacts, with low and high values affecting predictions differently.

The plot in figure 10 highlights how different IMU metrics contribute variably to predicting FOG, offering insights into feature importance and enabling interpretability of the model's decisions.

After processing this data, the model network uses the input features to predict the likelihood of Parkinson's disease. Transparency is promoted and the interpretation of the predicted outcomes is made easier by the application's clear presentation of the results as shown in fig.3. predict results button is used for making prediction. The models file format proves pivotal in preserving the model's integrity. The authors guarantee the preservation of the architecture, configuration, and trained parameters by encapsulating the model into the microntroller. This facilitates sharing, collaboration, and reproducibility in the field of Parkinson's disease prediction.

**Table 5:** Performance comparison for models:

| Metric | XGBoost | CatBoost | Extra Trees |
|---|---|---|---|
| **Precision (0)** | 0.93 | 0.96 | 0.93 |
| **Recall (0)** | 0.86 | 0.96 | 0.86 |
| **F1-Score (0)** | 0.89 | 0.96 | 0.89 |
| **Support (0)** | 2903 | 2903 | 2903 |
| **Precision (1)** | 0.88 | 0.96 | 0.88 |
| **Recall (1)** | 0.94 | 0.97 | 0.94 |
| **F1-Score (1)** | 0.91 | 0.96 | 0.91 |
| **Support (1)** | 3241 | 3241 | 3241 |
| **Accuracy** | 0.90 | 0.96 | 0.90 |
| **Macro Avg Precision** | 0.91 | 0.96 | 0.91 |
| **Macro Avg Recall** | 0.90 | 0.96 | 0.90 |
| **Macro Avg F1-Score** | 0.90 | 0.96 | 0.90 |
| **Weighted Avg Precision** | 0.90 | 0.96 | 0.90 |
| **Weighted Avg Recall** | 0.90 | 0.96 | 0.90 |
| **Weighted Avg F1-Score** | 0.90 | 0.96 | 0.90 |
| **Total Support** | 6144 | 6144 | 6144 |

A binary classification task was used to examine the performance characteristics of three machine learning models: XGBoost, CatBoost, and Extra Trees. Precision, recall, F1-score, and accuracy are computed for both classes (0 and 1) and averaged over the dataset as part of the evaluation.

In every statistic, CatBoost fared better than the other models. With the highest precision (0.96), recall (0.96), and F1-score (0.96) for class 0, CatBoost demonstrated exceptional performance in accurately detecting non-positive cases. CatBoost also demonstrated superior precision (0.96), recall (0.97), and F1-score (0.96) for class 1, demonstrating its efficacy in identifying affirmative cases. However, for classes 0 and 1, respectively, XGBoost and Extra Trees exhibited somewhat poorer precision (0.93 and 0.88), recall (0.86 and 0.94), and F1-scores (0.89 and 0.91).

With a total accuracy of 96%, CatBoost also outperformed XGBoost and Extra Trees, which both attained 90%. The precision, recall, and F1-scores, both macro-averaged and weighted-averaged, further highlight CatBoost's steady improvement (0.96) over XGBoost and Extra Trees (0.90). To ensure a fair comparison, the total support (6144 samples) stayed consistent across models. CatBoost is the most dependable model for this dataset, according to these findings. The model hyperparameters are tabulated in table 5.

**Table 5:** Hyperparameters for model

| Model | Accuracy |
|---|---|
| XGBoost | {'colsample_bytree': 1.0, 'learning_rate': 0.3, 'max_dept h': 6, 'n_estimators': 290, 'subsample': 1.0} |
| | |
| Catboost | 'depth': 8, 'iterations': 274, 'learning_rate': 0.3, 'subsample': 0.7544804990324112} |
| Extra Trees | {'max_depth': 20, 'max_features': 1.0, 'min_samples_leaf': 1, 'min_samples_split': 10, 'n_estimators': 402} |

**3.5 GPS module-based communication and tracking:**

The location tracking for the gait victim can be easily done, as seen in figure 11 (a). This allows users to effectively identify and track the location of the gait-affected person and provide aid to the victim immediately on need. The data retrieved via LoRa-based technology allows the users to easily track and aid the affected victim in any condition of detection of gait. The location tracking effectively supports the immediate aid needed in the case of emergencies using IoT platforms like Blynk or Python-based libraries that allow map-based tracking, and this can be delivered as an alert notification using SMTP libraries in Python, as shown in figure 11(b). This conceptual notification alert requires the use of any portable microcontroller, like Raspberry Pi, for better processing with the help of algorithm detailed below.

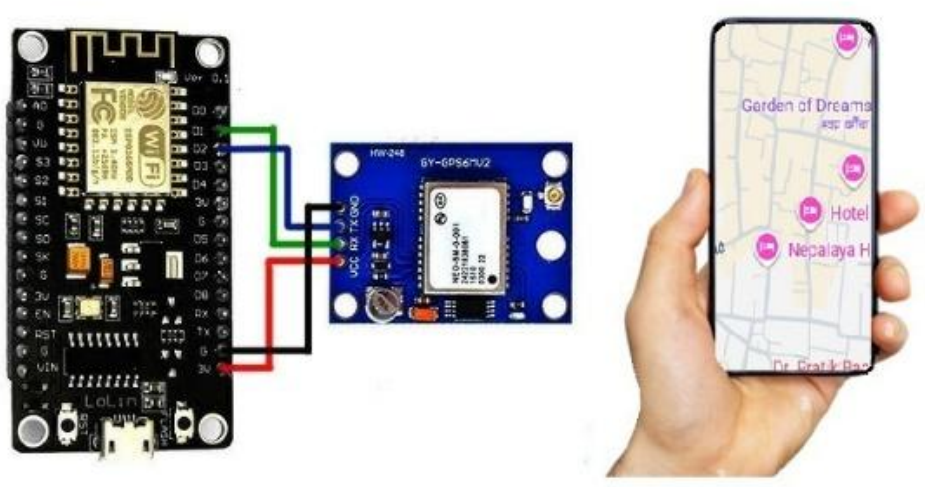

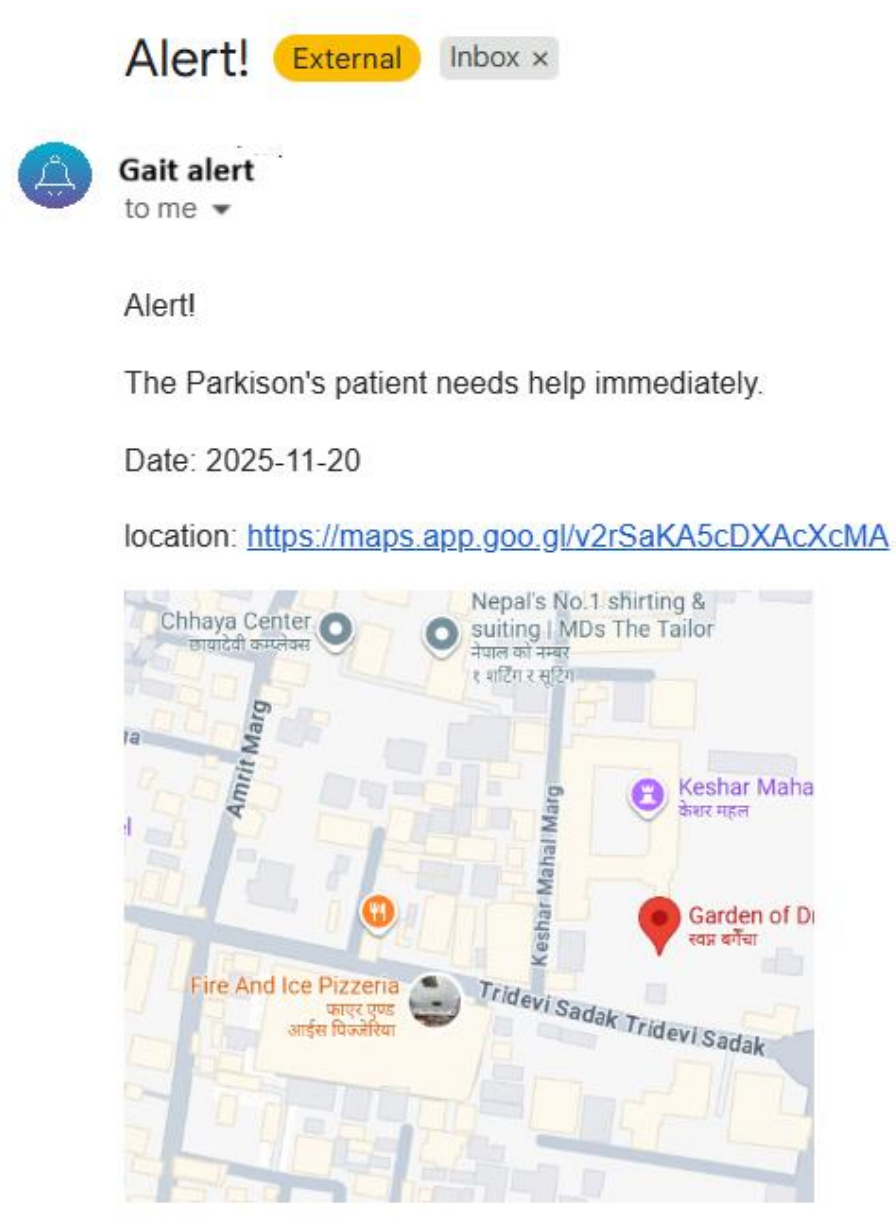

Fig 11 (a): Real-Time location tracking from GPS module Fig 11 (b): email alert

### 3.6 Algorithm (Step-by-Step):

1. *Receive GPS coordinates*

   *Get latitude (lat) and longitude (lon) from a GPS device or API.*

2. *Generate a map URL*

   *Use Google Maps Static API (or OpenStreetMap) to generate a map image URL based on the coordinates.*

3. *Download the map image*

   *Use Python's requests module to fetch the image from the URL.*

4. *Save the image locally*

   *Store the downloaded image as a .png or .jpg file.*

5. *Send the image*

   *Use Python (email, Telegram, WhatsApp, etc.) to send the saved image.*

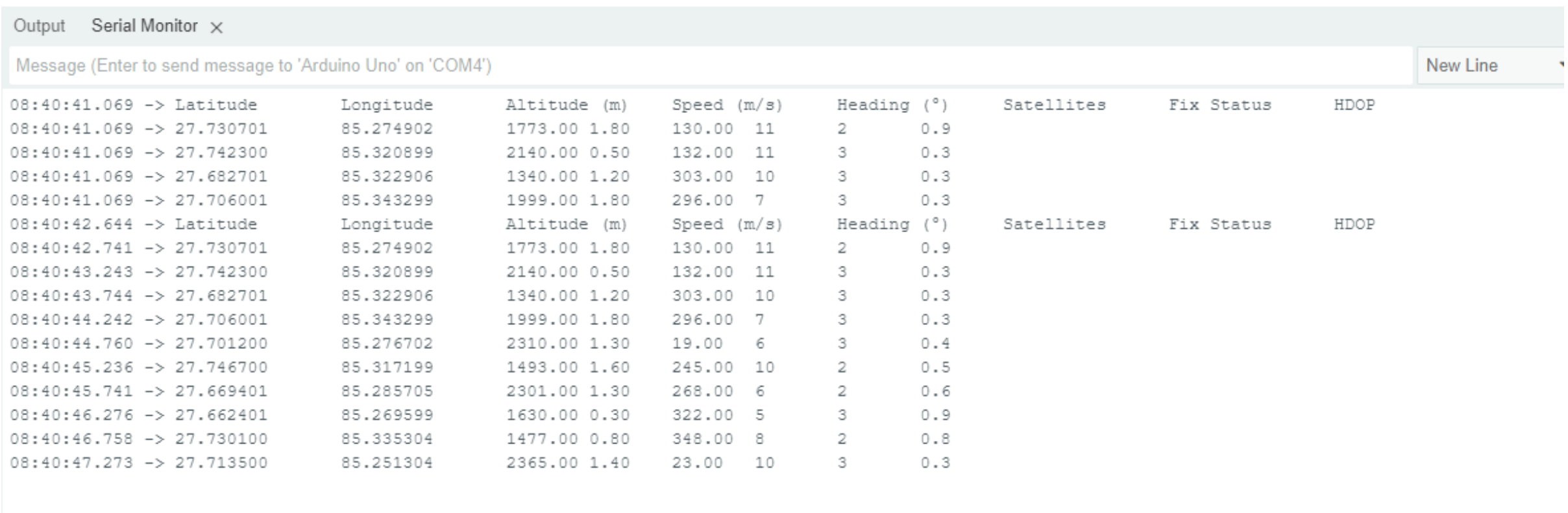

```
Output   Serial Monitor ×
Message (Enter to send message to 'Arduino Uno' on 'COM4')                New Line
08:40:41.069 -> Latitude     Longitude   Altitude (m)   Speed (m/s)   Heading (°)   Satellites   Fix Status   HDOP
08:40:41.069 -> 27.730701    85.274902   1773.00 1.80   130.00  11    2      0.9
08:40:41.069 -> 27.742300    85.320899   2140.00 0.50   132.00  11    3      0.3
08:40:41.069 -> 27.682701    85.322906   1340.00 1.20   303.00  10    3      0.3
08:40:41.069 -> 27.706001    85.343299   1999.00 1.80   296.00  7     3      0.3
08:40:42.644 -> Latitude     Longitude   Altitude (m)   Speed (m/s)   Heading (°)   Satellites   Fix Status   HDOP
08:40:42.741 -> 27.730701    85.274902   1773.00 1.80   130.00  11    2      0.9
08:40:43.243 -> 27.742300    85.320899   2140.00 0.50   132.00  11    3      0.3
08:40:43.744 -> 27.682701    85.322906   1340.00 1.20   303.00  10    3      0.3
08:40:44.242 -> 27.706001    85.343299   1999.00 1.80   296.00  7     3      0.3
08:40:44.760 -> 27.701200    85.276702   2310.00 1.30   19.00   6     3      0.4
08:40:45.236 -> 27.746700    85.317199   1493.00 1.60   245.00  10    2      0.5
08:40:45.741 -> 27.669401    85.285705   2301.00 1.30   268.00  6     2      0.6
08:40:46.276 -> 27.662401    85.269599   1630.00 0.30   322.00  5     3      0.9
08:40:46.758 -> 27.730100    85.335304   1477.00 0.80   348.00  8     2      0.8
08:40:47.273 -> 27.713500    85.251304   2365.00 1.40   23.00   10    3      0.3
```

Fig 12 (a): Serial monitor data for GPS

The GPS module serial data obtained through microcontroller is given in figure 12(a) and 12(b). Utilizing blynk technology alert message can be received successfully and finally, users can live monitor the victim or affected person's location for providing them the required emergency aid in the case of such hazardous situation for their health. The GPS module results as seen in the serial monitor shows the possibility of usage of such sensor-based approach for real time problem solving and detecting gait early in the process of assisting the victims. The LoRa receiver makes connectivity to the IoT system through receiving microcontroller and to finally, provide the assistive aid needed for the victim easily can be identified.

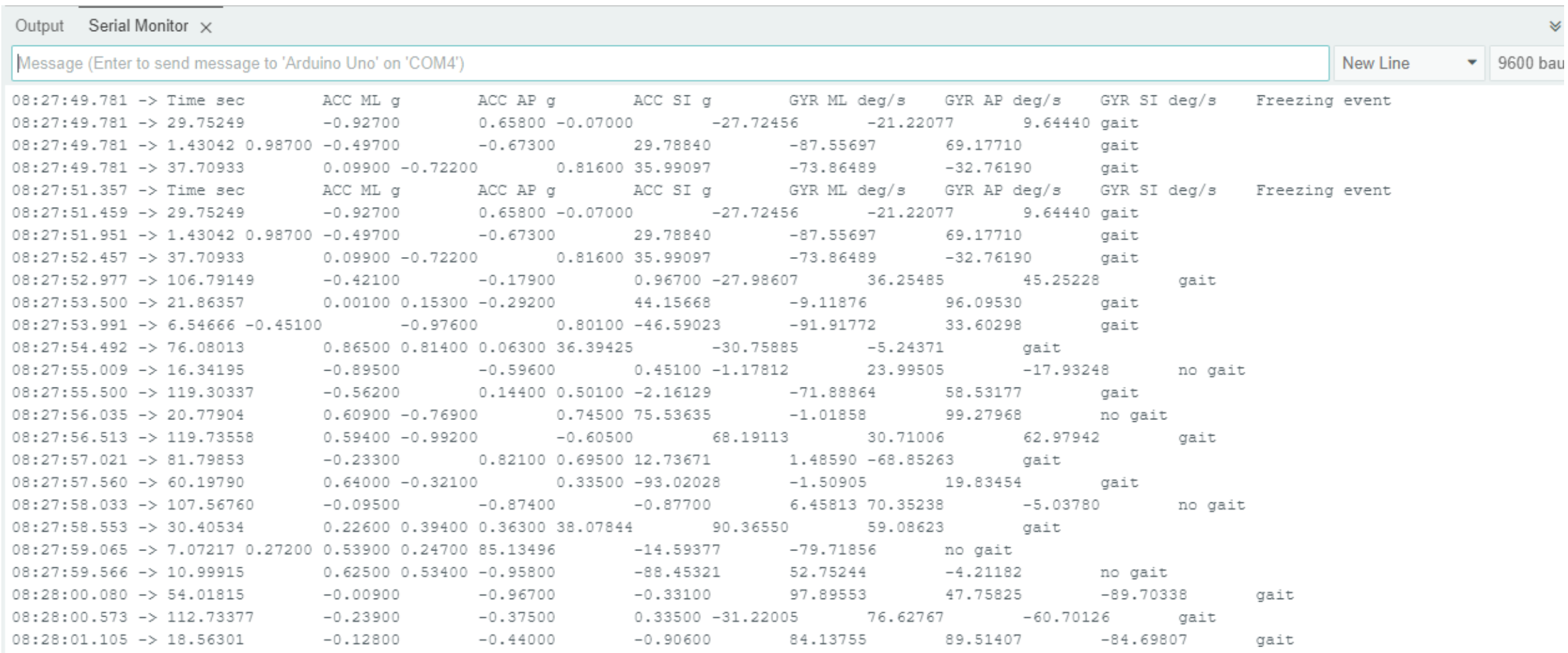

Fig 12(b) GPS various parameters in serial monitor

The monitoring on serial monitor shows the sensors data incoming through the sensors. The value 1, 2, 4 and others shows the various IMU sensors incoming data. This is received by model to make predictions for conditions related to gait or no gait detections. The serial plot result can be seen in figure 13 that shows various IMU gyro and accelerometer data variation over time.

This system shown on figure 12 (c) referenced from [21] uses WiFi via ESP8266 to transmit location data in real time. The system utilizes a Quectel L86 GPS module (or alternatives like Neo-6M) to capture latitude, longitude, speed, and bearing, which are then displayed on the Blynk dashboard with Google Maps integration. The GPS tracker operates using the trilateration method, connecting to multiple GNSS satellites to determine precise location data. The configuration required ESP8266, L86 GPS module, jumper wires, and breadboard, along with instructions for connecting the GPS module to the NodeMCU via serial communication. This demonstrates a low-power, compact, and real-time vehicle or asset tracking solution suitable for IoT applications, wearable devices, personal tracking, and use in emergency wearable sensors development.

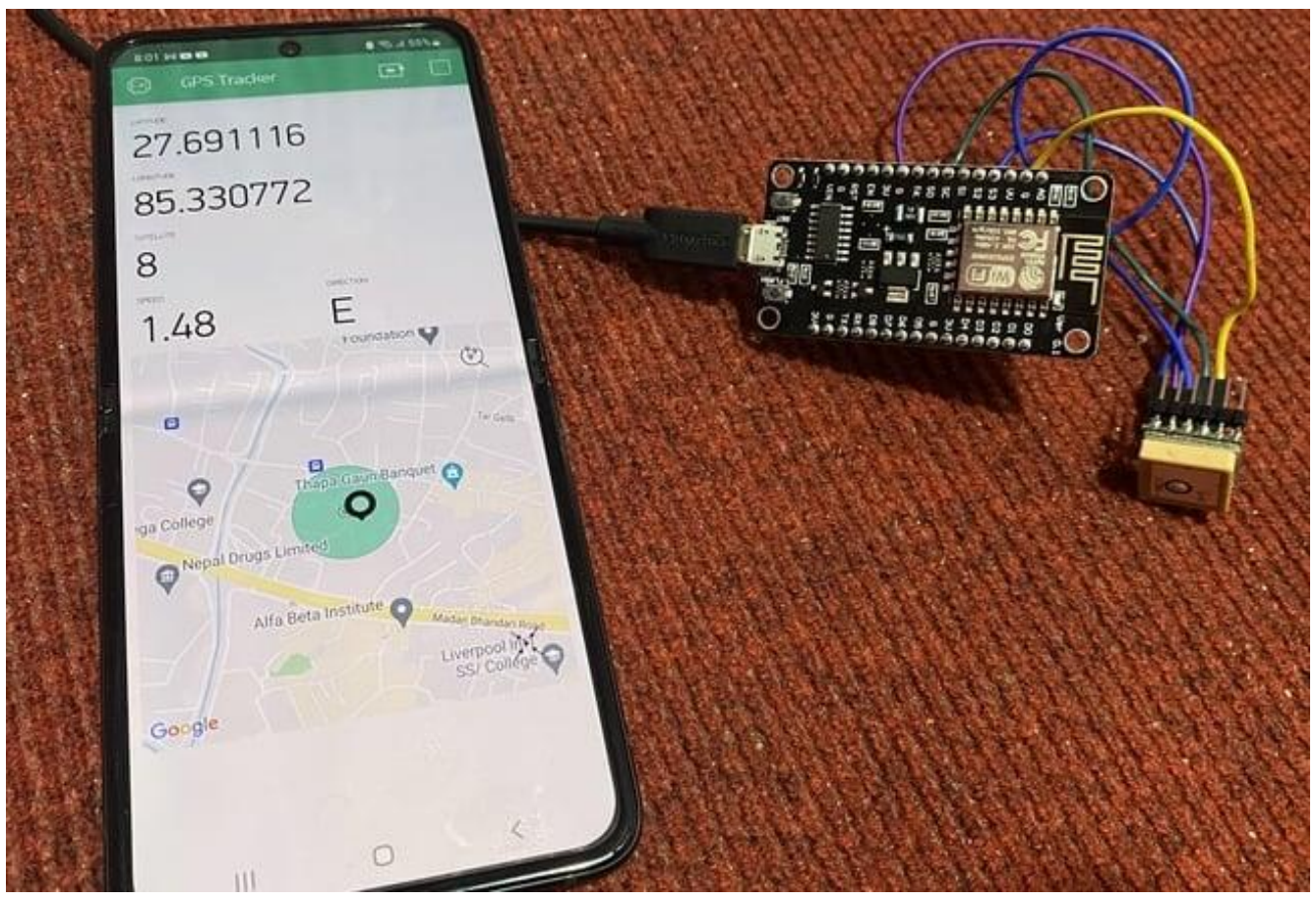

Fig 12 ( c): GPS tracking on blynk IoT [21]

**Table 6:** Sensor based data and prediction through model integrations

| | | | | | | | | |
|---|---|---|---|---|---|---|---|---|
| 08:28:45.791 | 61.75977 | 0.94600 | -0.33900 | 0.47800 | -71.74937 | -64.28495 | 8.78968 | gait |
| 08:28:46.308 | 80.07323 | 0.67700 | 0.39200 | -0.04700 | 78.51853 | -4.91729 | -24.59796 | no gait |
| 08:28:46.797 | 99.08819 | 0.53100 | -0.96500 | 0.46800 | -31.47390 | -65.32730 | 78.06266 | gait |
| 08:28:47.842 | 61.17399 | 0.66400 | -0.90700 | 0.88600 | -66.55076 | -52.99806 | 30.36767 | gait |
| 08:28:48.342 | 95.31682 | -0.19500 | -0.68600 | 0.65200 | -58.26415 | 91.97806 | 60.17960 | no gait |

The sensor data can be seen in table 6 and similarly, LoRa module serial data can be visualized in serial monitor as seen in figure 14. The test results on various iterations shows the model compatibility to predict the gait. The conditions helps to figure out whether the person is going to be attacked or not. The predictions that are made by model are tabulated in the table. The results show that the accurate prediction of gait allows use of sensor and IoT technology integration for reducing the threats related to freezing of gaits. The GPS data like latitude, longitude can be monitored through Esp-32 on serial monitor as seen in figure which shows at various time how the result from sensors progresses. The use of GPS module provides the concept for detecting and diagnosis of gait by accurately locating the person current locations.

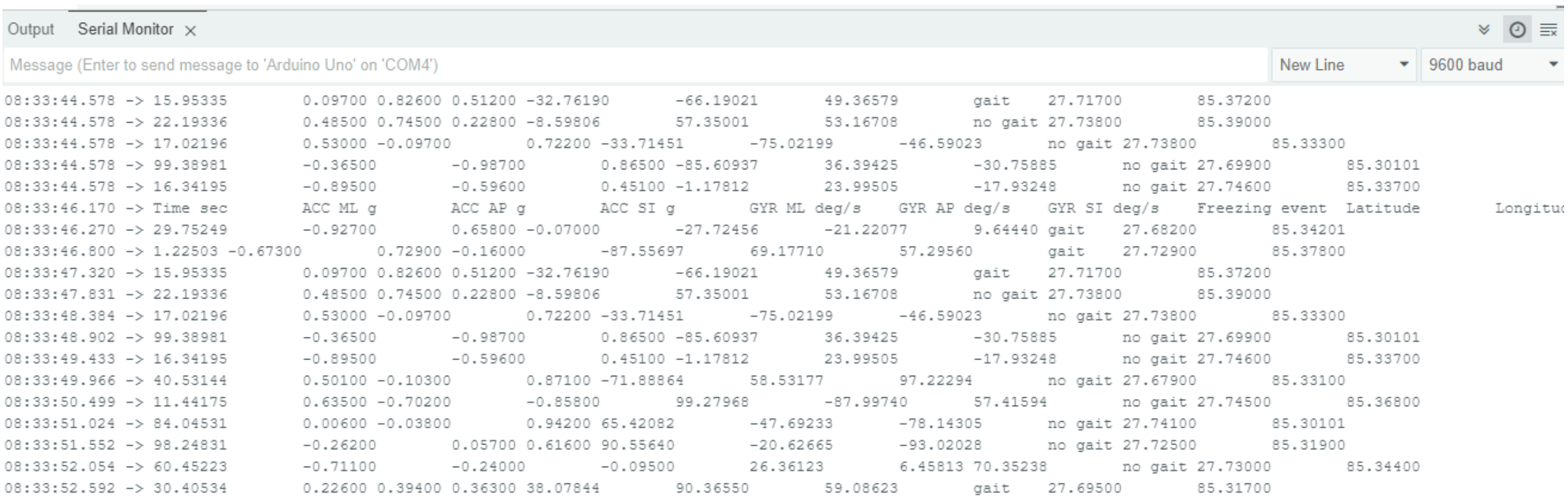

Fig 14: GPS data obtained using LoRa receiver side

The result shows a great potential of aiding the victim in the case of emergencies related to the occurrence of freezing of gait. The gait can be reduced utilizing such low-cost wearable sensor technology and tracked easily in emergency situations. The use of AI and IoT enhances the possibility of AI-powered ecosystem development as well as live monitoring, providing aid on time to the victim. The use of AI for biomedical prototyping and predictions and early detections [12, 13] can have significant enhancements in reducing health and various other emergencies. By use of such technologies, easily many kinds of diseases, apart from Parkinson's, can be healed [1, 2]. The use of sensor technologies has shown great possibilities in similar fields like biomedical assistive technology development [12] and enhancements by use of AI-powered tools and techniques.

**3.7 Limitations and Future works**

While there has been a greater advantage by use of machine learning for gait prediction, yet there is doubt on integrating with sensors for real-time prediction as the sensor data must align to the model predictability. The data received from sensors varied with time, and so it was difficult to correctly predict and identify gait disease. While the sensors are chosen for testing and calibration, a major disadvantage of using the MPU 6050 IMU sensor lies in the errors present on it, and the use of Kalman filtering somehow helped reduce those errors.

In the future, the use of extensive technologies for monitoring and tracking would be a greater advantage for the users to effectively aid the victims affected by the Freezing of Gait disease. The use of computer vision and highly precise sensing technologies would be more appropriate to enhance the system, and finally, using a higher-quality sensor like BNO-055 replacing the IMU MPU 6050 may provide a broader aspect for quality data retrieval for predicting gait easily.

## 4. Conclusions:

This study concludes by showing that classifying Gait using machine learning techniques is feasible. Although conventional machine learning algorithms perform admirably, deep neural networks are more adept at deciphering intricate relationships found in the data. With the XGBoost model, an excellent outcome was attained. The results highlight machine learning's potential to support early gait prediction and diagnosis. In order to increase predictive accuracy in actual clinical scenarios, future research could concentrate on growing datasets and improving model architectures.

In conclusion, the trade-offs between complexity and performance are reflected in the varying accuracy obtained across models. Even though models showed excellent accuracy, there is still a lot of potential for model and system improvement. Further highlighting the models' advantages and shortcomings, the confusion matrix's nuanced insights open the door to future machine learning advancements in Parkinson's disease classification.

**Declaration of clinical and ethical standard practice:**

Not applicable (No human subjects were used for experiments). Data from publication is adapted to train models

**Declaration of funding availability:**

No funding has been received for this work.

**Data availability statement:**

Data is publicly available through frontiers publication [17].

**Declaration of preprint availability statement:**

A preprint version of this article is published on arxiv:

https://doi.org/10.48550/arXiv.2507.01068

## REFERENCES

[1] Alshammri R, Alharbi G, Alharbi E, Almubark I. Machine learning approaches to identify Parkinson's disease using voice signal features. *Front Artif Intell*. 2023;6:1084001. doi: 10.3389/frai.2023.1084001. PMID: 37056913; PMCID: PMC10086231.

[2] Chatterjee K, Kumar RP, Bandyopadhyay A, Swain S, Mallik S, Li A, Ray KP. PDD-ET: Parkinson's Disease Detection Using ML Ensemble Techniques and Customized Big Dataset. *Information*. 2023;14(9):502. doi: 10.3390/info14090502.

[3] Nithya M, Lalitha V, Paveethra K, Kumari S. Early Detection of Parkinson's Disease using Machine Learning & Image Processing. *2022 International Conference on Computer Communication and Informatics (ICCCI)*, Coimbatore, India; 2022. p. 1–4. doi: 10.1109/ICCCI54379.2022.9740961.

[4] S E, T C, R S VD. Prediction of Parkinson's disease using XGBoost. *2022 8th International Conference on Advanced Computing and Communication Systems (ICACCS)*, Coimbatore, India; 2022. p. 1769–1772. doi: 10.1109/ICACCS54159.2022.9785227.

[5] Schmid M, Sheikhi S, Kheirabadi MT. An Efficient Rotation Forest-Based Ensemble Approach for Predicting Severity of Parkinson's Disease. *J Healthcare Eng*. 2022;5524852. doi: 10.1155/2022/5524852.

[6] Uddin MZ. A wearable sensor-based activity prediction system to facilitate edge computing in smart healthcare system. *J Parallel Distrib Comput*. 2018;123:46–53. doi: 10.1016/j.jpdc.2018.08.010.

[7] Pardoel S, Kofman J, Nantel J, Lemaire ED. Wearable-Sensor-Based Detection and Prediction of Freezing of Gait in Parkinson’s Disease: A Review. *Sensors*. 2019;19:5141. doi: 10.3390/s19235141.

[8] Mishra RK, Nunes AS, Enriquez A, et al. At-home wearable-based monitoring predicts clinical measures and biological biomarkers of disease severity in Friedreich’s Ataxia. *Commun Med*. 2024;4:217. doi: 10.1038/s43856-024-00653-1.

[9] Yue P, Li Z, Zhou M, Wang X, Yang P. Wearable-Sensor-Based Weakly Supervised Parkinson’s Disease Assessment with Data Augmentation. *Sensors*. 2024;24:1196. doi: 10.3390/s24041196.

[10] Pradhan MR, Ateeq K, Mago B. Wearable Device Based on IoT in the Healthcare System for Disease Detection and Symptom Recognition. *Int J Artif Intell Tools*. 2021;30(6–8):Article 2140011. doi: 10.1142/S021821302140011X.

[11] Habib Z, Mughal M, Khan M, Shabaz M. WiFOG: Integrating deep learning and hybrid feature selection for accurate freezing of gait detection. *Alexandria Eng J*. 2023;86:481–493. doi: 10.1016/j.aej.2023.11.075.

[12] Bikias T, Iakovakis D, Hadjidimitriou S, Charisis V, Hadjileontiadis LJ. DEEPFOG: An IMU-Based detection of freezing of GAIT episodes in Parkinson’s disease patients via Deep Learning. *Front Robot AI*. 2021;8:537384. doi: 10.3389/frobt.2021.537384.

[13] Borzì L, Sigcha L, Rodríguez-Martín D, Olmo G. Real-time detection of freezing of gait in Parkinson’s disease using multi-head convolutional neural networks and a single inertial sensor. *Artif Intell Med*. 2022;135:102459. doi: 10.1016/j.artmed.2022.102459.

[14] Salomon A, Gazit E, Ginis P, et al. A machine learning contest enhances automated freezing of gait detection and reveals time-of-day effects. *Nat Commun*. 2024;15:4853. doi: 10.1038/s41467-024-49027-0.

[15] Shi B, Tay A, Au WL, Tan DML, Chia NSY, Yen SC. Detection of Freezing of Gait Using Convolutional Neural Networks and Data From Lower Limb Motion Sensors. *IEEE Trans Biomed Eng*. 2022;69(7):2256–2267. doi: 10.1109/TBME.2022.3140258.

[16] Kim J, Porciuncula F, Yang HD, et al. Soft robotic apparel to avert freezing of gait in Parkinson’s disease. *Nat Med*. 2024;30:177–185. doi: 10.1038/s41591-023-02731-8.

[17] De Souza CR, Miao R, De Oliveira JA, et al. A public data set of videos, inertial measurement unit, and clinical scales of freezing of GAIT in individuals with Parkinson's disease during a Turning-In-Place task. *Front Neurosci*. 2022;16:832463. doi: 10.3389/fnins.2022.832463.

[18] Powers, D. M. (2020). Evaluation: from precision, recall and F-measure to ROC, informedness, markedness and correlation. arXiv preprint arXiv:2010.16061.

[19] Yang, W., Hamilton, J.L., Kopil, C. *et al.* Current and projected future economic burden of Parkinson's disease in the U.S.. *npj Parkinsons Dis.* **6**, 15 (2020). https://doi.org/10.1038/s41531-020-0117-1

[20] Towards Healthcare. (2025, March 24). *Parkinson's disease diagnosis and treatment market to hit $16.14 Bn by 2034*. Retrieved from https://www.towardshealthcare.com/insights/parkinsons-disease-diagnosis-and-treatment-market-sizing

[21] How2Electronics. (2022, May 11). *Real Time GPS Tracker using ESP8266 & Blynk with Maps*. How2Electronics. https://how2electronics.com/real-time-gps-tracker-using-esp8266-blynk-with-maps/